 \documentclass[pmlr,twocolumn,10pt]{jmlr} 

\usepackage{booktabs}
\usepackage{siunitx}

\usepackage{multirow} 
\usepackage{makecell} 
\usepackage{pdflscape} 

\usepackage[labelformat=simple]{subcaption}

\newcommand{\equal}[1]{{\hypersetup{linkcolor=black}\thanks{#1}}}

\theorembodyfont{\upshape}
\theoremheaderfont{\scshape}
\theorempostheader{:}
\theoremsep{\newline}

\jmlrvolume{LEAVE UNSET}
\jmlryear{2023}
\jmlrsubmitted{LEAVE UNSET}
\jmlrpublished{LEAVE UNSET}
\jmlrworkshop{Machine Learning for Health (ML4H) 2023} 

\title[Diagnosis Pathway Extraction]{Extracting Diagnosis Pathways from Electronic Health Records using Deep Reinforcement Learning}

\author{%
\Name{Lillian Muyama} \Email{lillian.muyama@inria.fr}\\
\addr Inria Paris, Paris, France\\ Centre de Recherche des Cordeliers, Inserm, Université Paris Cité, \\
Sorbonne Université, Paris, France
\AND
\Name{Antoine Neuraz}\equal{These authors contributed equally}  \Email{antoine.neuraz@aphp.fr}\\
\addr Hôpital Necker, Assistance Publique - Hôpitaux de Paris, Paris, France \\ Centre de Recherche des Cordeliers, Inserm, Université Paris Cité, \\
Sorbonne Université, Paris, France
\AND
\Name{Adrien Coulet}\footnotemark[1] \Email{adrien.coulet@inria.fr}\\
\addr Inria Paris, Paris, France\\ Centre de Recherche des Cordeliers, Inserm, Université Paris Cité, \\
Sorbonne Université, Paris, France
}

\begin{document}

\maketitle

\begin{abstract}
Clinical diagnosis guidelines aim at specifying the steps that may lead to a diagnosis. 
Inspired by guidelines, we aim to learn the optimal sequence of actions to perform in order to obtain a correct diagnosis from electronic health records. We apply various deep reinforcement learning algorithms to this task and experiment on a synthetic but realistic dataset to differentially diagnose anemia and its subtypes and particularly evaluate the robustness of various approaches to noise and missing data.
Experimental results show that the deep reinforcement learning algorithms show competitive performance compared to the state-of-the-art methods with the added advantage that 
they enable the progressive generation of a pathway to the suggested diagnosis, which can both guide and explain the decision process.
\end{abstract}

\section{Introduction} \label{sec:introduction}
Clinical diagnosis guidelines are documents to guide, rationalize and normalize clinical decisions, classically established by a college of experts on the basis of the best available evidence \citep{field1990clinical}.
They mainly describe the steps that may lead to a diagnosis, such as information collection, observations, and laboratory test orders. 
However, clinical guidelines suffer several drawbacks. First, they are designed to cover the majority of the population and hence may fail to guide to the right diagnosis in the case of uncommon patients such as those with multiple diseases. Second, their establishment is long and expensive, and updating them is usually done after several years \citep{steinberg2011clinical}. This makes them unsuitable to fast emerging practices such as those associated with a recently developed laboratory test or an emerging disease. Moreover, as clinical guidelines are expensive and time-consuming to produce, this development approach does not scale to the full spectrum of diseases. Thus, more versatile and scalable methods are required to provide insights when clinical guidelines are not available.

We think that machine learning approaches trained on clinical data may complement diagnosis guidelines. In particular, we aim at developing approaches able to guide each step of the decision process, as described in \citet{adler2021}. We believe that such an approach may reduce the number of irrelevant tests therefore optimizing healthcare costs, but primarily may propose more personalized and accurate diagnoses, especially in the case of patients with uncommon conditions. 

The collection of patient-level data in Electronic Health Records (EHRs) offers great opportunities to gain knowledge about clinical practice \citep{jensen2012}. 
EHRs encompass structured, semi-structured and unstructured data about patients' health such as medications, laboratory test orders and results, diagnoses, as well as demographic features. Previous works have trained machine learning (ML) methods on EHRs to automatically suggest diagnoses for patients such as in \citet{lipton2015learning}, \citet{miotto2016} and \citet{choi2016doctor}. However, in these studies, supervised ML methods are employed to predict a unique endpoint, \textit{i.e.}, the diagnosis represented as a class label.
We believe that for data-driven approaches to find adoption in clinical practice, it is important for a diagnosis not to be limited to an endpoint, but to be represented as a pathway that follows steps of medical reasoning and decision-making. 

In this work, we propose to leverage EHR data and investigate how it can be used to train a family of Reinforcement Learning (RL) methods to build explainable pathways for the differential diagnosis of anemia, as a primary use case. Anemia is a clinical condition defined as a lower-than-normal amount of healthy red blood cells in the body, which we chose for three reasons: its diagnosis is made mainly from a series of laboratory tests that are available in most EHRs; it is a common diagnosis implying that the associated amount of data may be sufficient to train RL models; and the differential diagnosis of anemia is frequently complex to establish, making its guidance useful.

We propose to use RL because it builds a model that is able to pass through various \emph{actions} and \emph{states} to reach a final objective state. We adapted the framework of RL to construct individualized pathways of observations in a step-by-step manner, in order to suggest a decision. In our particular use case, a pathway is a sequence of laboratory test requests (actions), whose results are then obtained (states) before either requesting for another test or terminating on an anemia differential diagnosis. 
We make the assumption that the constructed pathways can complement clinical guidelines to aid practitioners in decision-making during the diagnosis process. 

Our main contributions are:
\textit{(i)}
    an adaptation of the RL framework to progressively construct  
    optimal sequences of actions to perform in order to reach a diagnosis and
\textit{(ii)} 
    an empirical analysis that identifies the most suitable algorithm for our use-case, and 
    evaluates the robustness of our approach in regards to levels of missing and noisy data.

\section{Related Work} \label{sec:related_work}

Previously, studies have used various process mining, machine learning and statistical methods to extract clinical pathways from medical data. \citet{zhang2015paving} proposed Markov Chains for the identification of clinical pathways using patient visits data. \citet{perer2015mining} extracted common patterns in the data in order to build pathways while \citet{baker2017process} used a Markov model to extract meaningful clinical events from patient data. \citet{huang2013latent, huang2014discovery, huang2018probabilistic} took a more statistical approach to build treatment pathways by using a Latent Dirichlet Allocation-based method. 
Machine learning methods such as Long Short-Term Memory and reinforcement learning have also been used to construct pathways for optimal treatment in several use-cases such as in \citet{lin2021personalized} and \citet{li2022electronic} respectively. However, the focus of all these papers is building pathways for the \emph{treatment} of patients with a specific medical condition. In our paper, we aim to build pathways for the \emph{diagnosis} of medical conditions.

Likewise, numerous studies have leveraged machine learning methods for disease diagnosis. Given the longitudinal nature of Electronic Health Record (EHR) data, a prevalent choice in these investigations has been Recurrent Neural Networks (RNNs), as observed in the works of \citet{lipton2015learning} and \citet{choi2016doctor}. Additionally, RNNs have also been used in the prediction of future patient outcomes such as in \citet{koshimizu2020prediction}. In \citet{obaido2022interpretable}, \citet{zoabi2021machine} and \citet{ kavya2021machine}, they go further by not only employing ML approaches to provide clinical diagnoses for the patients, but also using explainable Artificial Intelligence (AI) methods to interpret the model results.
However, our aim is not only to diagnose the patient with the correct condition, but also to find the optimal sequence of features to test to reach this diagnosis for each patient. In other terms, our aim is to construct personalized diagnosis pathways that delineate the steps leading to the diagnosis, thereby explaining the diagnostic process. 

Previous works used reinforcement learning methods for costly feature acquisition in classification tasks such as in \citet{li2021active} and \citet{janisch2019classification}. However, these works did not primarily focus on the clinical diagnosis task. \citet{yu2023deep} aimed to optimize the financial cost of the three prediction tasks, including the prediction of Acute Kidney Injury, using RL. While it was shown that their method reduced the cost of diagnosis, the actions taken at each step in the diagnosis process were not shown, therefore the pathways' overall relevance and explainability were unknown. In addition, \citet{tang2016inquire}, \citet{wei2018task} and \citet{kao2018context} used RL to diagnose patients by inquiring about the presence of disease-related symptoms from users. While our approach similarly formulates the diagnosis process as a sequential decision-making problem and applies deep reinforcement learning (DRL) techniques, these studies use a symptom self-checking approach, whereas we aim to use EHR data. We believe that our approach is more suitable as EHRs encompass data collected routinely in clinical practice, including objective and normalized measurements such as laboratory results which we are proposing to use in this first study. 

\section{Methods} \label{sec:methods}
\subsection{Decision Problem}
We consider the anemia diagnosis process as a sequential decision-making problem and formulate it as a Markov Decision Process (MDP) \citep{Littman2001markov}, following the RL \citep{sutton2018reinforcement} paradigm. 
Accordingly, we define an \emph{agent} interacting with an \emph{environment} in order to maximize a cumulative reward signal. 
At each time step $t$, the agent receives an observation $o$ from the environment \emph{state}, takes an \emph{action}, and obtains a \emph{reward}.  
The goal is to learn a policy, \textit{i.e.}, a function that maps states to actions, that maximizes the reward signal. 
In this study, our agents are taking observations from the synthetic EHRs of a Clinical Data Warehouse (CDW), 
in order to reach a diagnosis, which is a final action. 
Let $\mathcal{D}$ denotes a dataset of size $n \times (m+1)$, associated with $F$, the set of names of the $m$ features, and $C$ the set of possibles diagnosis values. An instance $D^i$ in $\mathcal{D}$ is a pair $(X^i, Y^i)$ with $X^i = \{x^i_1,\dots, x^i_j, \dots , x^i_m\}$ where $x^i_j$ is the value of the feature $j$ for the patient $i$, $m$ is the total number of features and $Y^i \in C$ is the anemia diagnosis of patient $i$.
Accordingly, our MDP is defined by the quadruple  $(S, A, T, R)$  as follows:

\begin{itemize}
    \item $S$ is the set of states. 
    At each timestep, the agent receives an observation $o$ of the state $s_t$, which is a vector of fixed size $m$ comprising the values of the features that have already been queried by the agent at time $t$; 
    features that have not been queried yet are associated with the value -1. Equation \ref{eqn:state_eqn} 
     defines the $j^{th}$ element of $o$, where $F'$ denotes the set of features that have already been queried by the agent. 
    \begin{equation}
    \label{eqn:state_eqn}
      o_j = \begin{cases}
      \;x_j, & \text{if $f_j \in F' $}\\
      -1, & \text{otherwise.}
    \end{cases}
    \end{equation}
    
    \item $A$ is the set of possible actions, which is the union of the set of \emph{feature value acquisition actions}, $A_f$ (or value actions for short),  and the set of 
    \emph{diagnostic actions}, $A_d$. 
     At each time step, the agent takes one action $a_t \in A$. 
     Actions from $A_f$ are taking values from the set of features $F$. Accordingly a specific value $f_j \in F$ will trigger the action of querying for the value of this feature from the CDW.
     Actions from $A_d$ are taking values from the set of possible diagnoses $C$. 
     At a time step, if $a_t \in A_d$, the episode is terminated. Also, if an episode reaches the number of maximum specified steps without reaching a diagnosis, it is terminated.
    
    \item $T$ denotes the transition function that gives the probability of moving from a state $s_t$ to a state $s_{t+1}$ given an agent action $a_t \in A$.
    
    \item $R$ is the reward function. $r_{t+1}$, which can be written as $R(s_t, a_t)$, is the immediate reward when an agent takes an action $a_t$ in a state $s_t$. In the case of a diagnostic action $a_t \in A_d$, if the diagnosis is correct, the reward is +1, 
    otherwise the reward is -1. For a value action $a_t \in A_f$, if the feature has already been queried, the agent receives a reward of -1 and the episode is terminated. 
    Otherwise, the agent receives a reward of 0. Accordingly, the reward functions for
    diagnostic and value actions, are formalized as follows: 
     \begin{equation}
    \label{eqn:diag_actions}
       \text{if}\ a_t \in A_d, R(s_t, a_t) = \begin{cases}
      \;\;\,1, & \text{if $a_t = Y^i $}\\
      -1, & \text{otherwise}
    \end{cases}
    \end{equation}
    
     \begin{equation}
     \label{eqn:feat_actions}
      \text{if}\ a_t \in A_f, R(s_t, a_t) = \begin{cases}
      -1, & \text{if $a_t \in F' $}\\
      \;\;\,0, & \text{otherwise}.
    \end{cases}
    \end{equation}

\end{itemize}


\subsection{Q-learning and Extensions}

\textbf{Q-learning} \citep{watkins1992q} is an RL algorithm that outputs the best action to take in a given state based on the expected future reward of taking that action in that particular state. The expected future reward is named the Q-value of that state-action pair,
noted $Q(s,a)$. At each time step, the agent selects an action following a policy $\pi$, and the goal is to find the optimal policy $\pi^*$ that maximizes the reward function. 
During model training, the Q-values are updated using the Bellman Equation as described in Appendix \ref{apd:Q}, equation \ref{eqn:q_learning}.

In our use case, since the state space is large and continuous, we propose to use a \textbf{Deep Q-Network (DQN)} \citet{mnih2015human} which uses a neural network to approximate the Q-value function. 
In order to improve DQN stability and performance, several extensions of the DQN algorithm have been developed. 
Particularly \textbf{Double DQN (DDQN)}, \textbf{Dueling DQN} and \textbf{Prioritized Experience Replay (PER)}, which we use in the following of this paper and briefly describe in Appendix \ref{apd:Q}. 
These techniques enhance the performance of the DQN algorithm by improving its stability, making the learning process faster and more efficient, and reducing overestimation bias, among other benefits. 

\subsection{State-of-the-Art Classifiers}
As part of the study, we compared the performance of the DRL models with four classical supervised learning algorithms that are commonly used for classification tasks, namely Decision Tree (DT), Random Forest (RF), Support Vector Machine (SVM), and a Feed-Forward Neural Network (FFNN). 
DT has a particular status in our study for two reasons. First, we experimented on synthetic data, whose labels were assigned according to a decision tree. For this reason, the DT approach is expected to perform very well on our data. Second, DT is the only considered classifier that is self-explanatory as the path in the decision tree that is taken to classify an instance constitutes a pathway to the diagnosis.

\subsection{Dataset Synthesis}
In order to develop and evaluate the performance of our various agents, we built a synthetic dataset. 
Firstly, we identified a set of relevant variables from existing literature about the diagnosis of different anemia sub-types and after discussions with a domain expert. This concluded with the identification of 17 features and 8 anemia classes. 
Secondly, we constructed a dataset for each anemia class based on the decision tree adapted from the literature and presented in Figure \ref{fig:tree}. 
The values for each feature were generated using a uniform probability distribution. 
The final dataset comprises feature values for 70,000 patients.
A more detailed description of the dataset synthesis is in Appendix \ref{apd:dataset}. The class distribution is illustrated in Figure \ref{fig:class_distribution}; the ratio of observed \textit{vs.} missing values for each feature is shown in Figure \ref{fig:missing_values}; descriptive statistics of the dataset are shown in Table \ref{tab:statistics}; and an example of an instance from the dataset is provided Table \ref{tab:sample}.

Additionally, in order to compare the robustness of various approaches to imperfect data, we defined functions to artificially introduce different levels of noisiness and missingness to our training dataset as described in section \ref{sec:imperfect} of the Appendices.


\subsection{Evaluation Approach and Implementation}
80\% of the dataset was used to train the model while 20\% was used as the test set. Additionally, 10\% of the training dataset (8\% of the dataset) was used for validation. The validation and test sets used in all the experiments are constant \textit{i.e.}, without noise or missingness. Only training sets are changed. 

The primary metric used for evaluating the performance of our models is the \emph{accuracy}, which is the ratio of episodes that terminated with a correct diagnosis. We further computed the average number of actions performed for each episode, displayed as the \emph{mean episode length}. This is not applicable to the state-of-the-art classifiers since those consider all available features. The \emph{F1} and \emph{ROC-AUC} scores were also computed using a one-vs-rest approach and macro-averaging. Ten runs were conducted for the first experiment comparing DQN approaches and five runs for the subsequent experiments on the models' robustness. Each run was performed with the same datasets but with different seeds, which influence the models' interactions with the environment as well as their initial weights. 

We used the OpenAI Gym Python library \citep{brockman2016openai} to implement our environment. 
To build our agents, we used the stable-baselines package \citep{stable-baselines}, and the values of the hyperparameters are as shown in Table \ref{tab:hyperparams} in Appendix \ref{apd:params}. They were chosen through prior knowledge, existing literature, and experimentation. The hyperparameters for the SOTA models were chosen using a grid search strategy. The source code is available at \url{https://github.com/lilly-muyama/anemia_diagnosis_pathways}.

\section{Results} \label{sec:results} 
\subsection{Performance Comparison of DQN and its extensions}
In the first round of experiments, we trained DQN, Double DQN, Dueling DQN and Dueling Double DQN on our dataset; and also enabled prioritized experience replay (PER) for each of these models. We also trained a Proximal Policy Optimization (PPO) algorithm using a similar number of timesteps for comparison purposes. The associated results are compiled in Table \ref{tab:dqn_results}. Model names associated with the suffix -PER correspond to the mentioned DQN extension together with PER.
In order to gauge the stability, for each model, we conducted ten different runs of the same experiment using different seeds. The values in Table \ref{tab:dqn_results} are therefore averages and standard deviations (SD) over the ten runs. 
The number of time steps used was determined in the first run using the validation data and remained constant for the other nine runs.

Dueling DQN-PER provided the best accuracy and the second lowest SD, while Dueling DDQN-PER exhibited a similar performance with slightly lower accuracy and slightly less variance. For these reasons, we used only these two DQN models for the rest of the experiments.
\begin{table*}[!ht]
\centering
\floatconts
  {tab:dqn_results}
  {\caption{Performance of the RL models.}}%
  {%
    \begin{tabular}{|p{3.5cm}|p{2.5cm}|p{2.5cm}|p{2.6cm}|p{2.6cm}|}
    \hline
    \bfseries Model & \bfseries Accuracy & \bfseries Mean episode length & \bfseries F1 & \bfseries ROC-AUC\\\hline
         PPO & 25.575 $\pm$ 16.537  & 1.628 ± 0.852 & 18.422 ± 25.227 & 54.733 ± 16.172 \\ 
         DQN & 89.309 $\pm$ 14.684  & 4.272 ± 0.449 & 88.728 ± 15.946 & 93.904 ± 8.280\\
         DDQN & 92.792 $\pm$ 5.370 & 4.660 ± 0.312 & 92.183 ± 6.249 & 95.742 ± 3.278 \\ 
         Dueling DQN & 92.459 $\pm$ 5.313 & 4.697 ± 0.239 & 91.722 ± 6.568 & 95.645 ± 3.266 \\ 
         Dueling DDQN & 77.147 $\pm$ 31.987 & 4.444 ± 1.018 & 74.358 ± 34.975 & 86.842 ± 18.059 \\ 
         DQN-PER & 93.708 $\pm$ 5.509 & 4.673 ± 0.288 & 93.615 ± 5.195 & 96.412 ± 3.034  \\ 
         DDQN-PER & 95.571 $\pm$ 2.007  & 4.917 ± 0.355 & 95.336 ± 2.115 & 97.459 ± 1.163 \\ 
         \textbf{Dueling DQN-PER} & \textbf{96.639 $\pm$ 1.462} & \textbf{4.588 ± 0.201} & \textbf{96.498 ± 1.515} & \textbf{98.106 ± 0.823} \\
         Dueling DDQN-PER & 96.330$ \pm$ 1.316 & 4.821 ± 0.448 & 96.168 ± 1.351 & 97.863 ± 0.813 \\\hline
    \end{tabular}
  }
\end{table*}

\subsection{Performance Comparison with SOTA classifiers}
The results of these experiments are shown in Table \ref{tab:sota_results}. Since diagnosis labels of the synthetic dataset are assigned following the decision tree in Figure \ref{fig:tree}, the tree-based agent, which is based on the same tree achieved a perfect score. Also, we built an agent that acts randomly at each timestep for comparison purposes. Because the SOTA classifiers use a constant set of features to make a diagnosis, they do not have a mean episode length. The tree-based algorithms and the FFNN performed better than the DQN models, while SVM had a lower performance. 
\begin{table*}[!ht]
\centering
\floatconts
  {tab:sota_results}
  {\caption{Performance of the DQN and the state-of-the-art classifiers. The tree-based agent has a perfect score because it acts according to the decision tree used to build the dataset}}%
  {%
    \begin{tabular}{|p{3.2cm}|p{2.7cm}|p{2.5cm}|p{2.7cm}|p{2.7cm}|}
    \hline
    \bfseries Model & \bfseries Accuracy $\pm$ SD & \bfseries Mean episode length & \bfseries F1 & \bfseries ROC-AUC\\\hline
         Random Agent & 12.343 $\pm$ 0.331 & 1.534 ± 0.009 & 12.337 ± 0.326 & 49.996 ± 0.204 \\
         \textbf{Tree-based Agent} & \textbf{100.000 $\pm$ 0.000} & \textbf{3.975 ± 0.000} & \textbf{100.000 $\pm$ 0.000} & \textbf{100.000 $\pm$ 0.000}\\
         Decision Tree & 99.960 $\pm$ 0.004 & N/A & 99.961 ± 0.004 & 99.979 ± 0.002 \\
         Random Forest & 99.897 $\pm$ 0.012 & N/A & 99.898 ± 0.011 & 99.946 ± 0.006 \\
         FFNN & 97.966 $\pm$ 0.274 & N/A & 97.914 ± 0.281 & 98.806 ± 0.161 \\
         SVM & 94.893 $\pm$ 0.000 & N/A & 94.363 ± 0.000 & 96.795 ± 0.000 \\
         Dueling DQN-PER & 96.639 $\pm$ 1.462 & 4.588 ± 0.201 & 96.498 ± 1.515 & 98.106 ± 0.823 \\
         Dueling DDQN-PER & 96.330$ \pm$ 1.316 & 4.821 ± 0.448 & 96.168 ± 1.351 & 97.863 ± 0.813 \\\hline
    \end{tabular}
  }
\end{table*}

\subsection{Varying levels of missing data and noisiness}
\begin{figure*}[!ht]
\centering 
\floatconts
  {fig:results_plots}
  {\caption{Accuracy of approaches with varying levels of missingness, noisiness, and train set size. The graphs show the accuracy (median) of the models at different (\textit{a}) missingness levels; (\textit{b}) noisiness levels; (\textit{c}) missingness levels at a constant noisiness level (0.2). (\textit{d}) shows the accuracy (mean) and the 95\% confidence interval of the models based on the size of the train set.}}
  {%
    \subfigure{\label{missingness}%
      \includegraphics[width=0.45\linewidth]{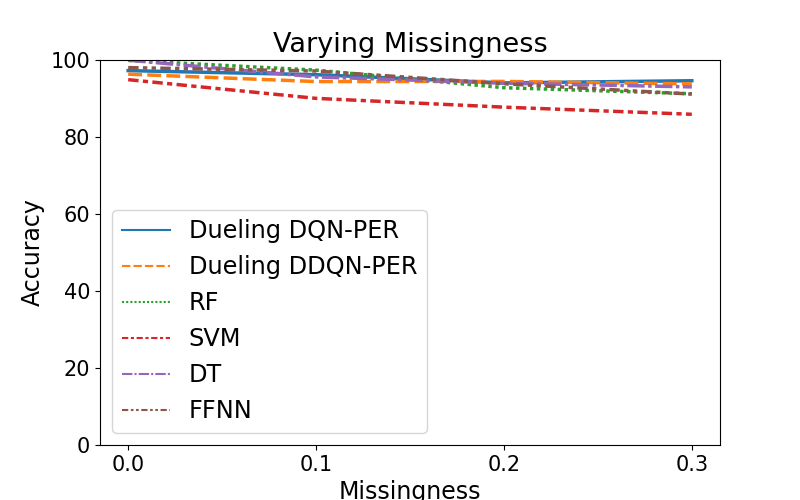}}%
    \qquad
    \subfigure{\label{noisiness}%
      \includegraphics[width=0.45\linewidth]{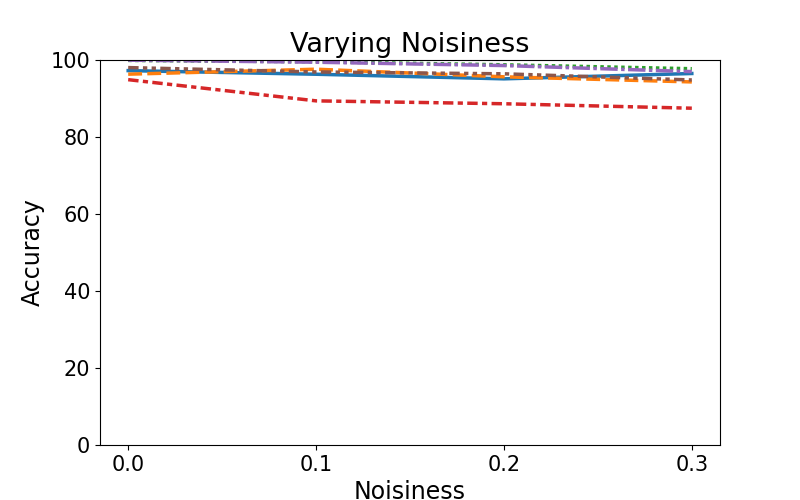}}
      
    \vspace{10pt} 

    \subfigure{\label{missingness_noisiness}%
      \includegraphics[width=0.45\linewidth]{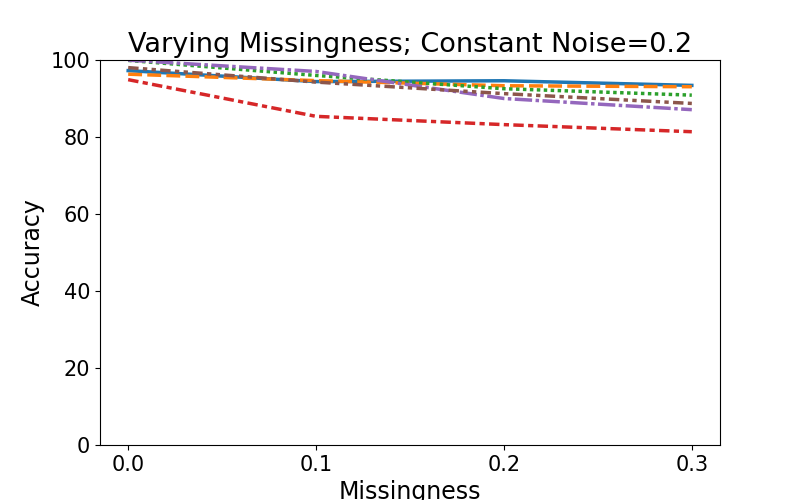}}%
    \qquad
    \subfigure{\label{varying_sizes}%
      \includegraphics[width=0.45\linewidth]{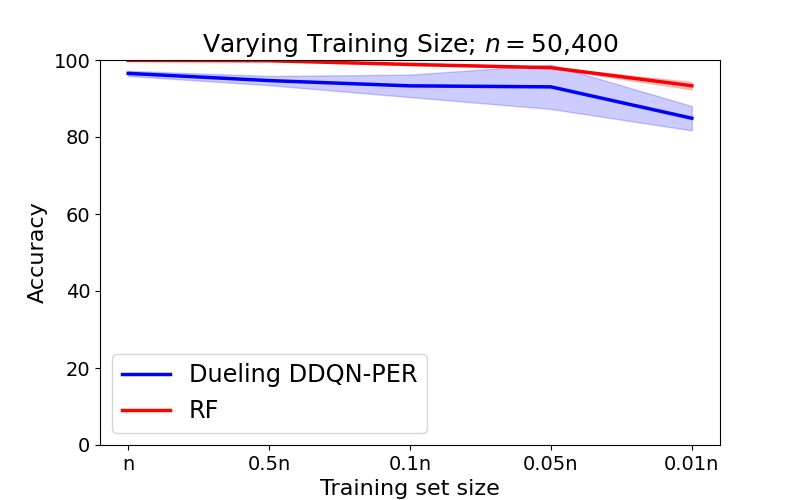}}
  }
\end{figure*}

 \figureref{missingness} depicts the accuracy when the level of missingness added to the training dataset varies. All the models' performance declined at a steady rate, with the DQN models' accuracy declining more slowly than the rest. SVM had the speediest deterioration. 
\figureref{noisiness} shows the accuracy when the level of noise in the training set varies. SVM showed the sharpest decline, while the rest of the models performed comparably, and the Dueling DQN-PER model exhibited a consistent performance across all noise levels.
\figureref{missingness_noisiness} is similar to \figureref{missingness} as it depicts the varying accuracy as the level of missingness increases but with a fixed level of noise of 0.2. Once more, SVM performance declined rapidly and the DT's performance decreased faster than the DQN models or the other classifiers. The two DQN models exhibited a consistent performance.
The depicted accuracy in Figures \ref{missingness}-\ref{missingness_noisiness} is the median of the accuracy as five runs were conducted for these experiments.

\subsection{Varying the size of train sets}
In order to understand whether this method can be applied to the diagnosis of conditions where data may be scarce, such as with rare diseases, we conducted experiments with train sets of different sizes. For each size, a fraction of the train set was randomly selected and removed, leaving the remaining data to form the new set. Runs were repeated five times for each size of the train set, whereby a different subset was removed each time. The Dueling DDQN-PER model was used for these experiments and for each run, the same number of timesteps was used. Figure \ref{varying_sizes} shows the mean accuracy for each train set size. The shaded region represents the 95\% confidence interval. The RF classifier is also depicted in the figure for comparison and it was chosen for its stability and high performance. As depicted, the DQN model was able to learn the clinical pathways and diagnose the patients. However, for all train set sizes, its performance was lower than the RF and as the size of the train set decreased, the model's performance became less stable.

\section{Discussion} 
As illustrated by our study, applying deep reinforcement learning methods to EHR data can be used to learn 
clinical diagnosis pathways for patients whereby at each step, a different action is taken by the agent until a diagnosis is reached. In comparing our two best DQN models 
to state-of-the-art classifiers, while the latter perform extremely well on perfect data, the two DQN models performed comparably and sometimes better on imperfect data, showing their robustness. 
Furthermore, using different training sizes, the DQN Models showed their capacity to learn clinical diagnosis pathways associated with decent performance, even in the context of reasonably small-sized datasets. 

\subsection{Generated Pathways}
Besides quantitative evaluation, the pathways constructed by DQN models present two qualities. The fact that they are composed of progressive observations that lead to a diagnostic decision, makes them explainable: One may understand why a particular diagnosis is reached by looking at the sequence of features selected and their associated values. In addition, the set of pathways generated by a test set can be aggregated to generate a data structure similar to diagnosis guidelines. We note that the DT algorithms are also intrinsically explainable, but our results show that they are less robust to imperfect data. Sample generated pathways are shown in Appendix \ref{apd:pathway}. Interactive plots of the pathways generated by the model can be accessed at \url{https://lilly-muyama.github.io/}.

\subsection{DQN Algorithms Performance}
From Table \ref{tab:dqn_results}, DDQN and Dueling DQN exhibit similar performance and illustrate that both extensions improve the accuracy and stability of the standard DQN. 
However, Dueling DDQN exhibited the lowest performance of all the algorithms and showed wide variability. This may be because both Double DQN and Dueling DQN were created to solve the same problem: the overestimation of Q-values. Therefore, using both techniques at the same time may not lead to better performance as long as the Q-value is still estimated using the same technique as in Double DQN. Additionally, using both Double DQN and Dueling DQN at the same time makes the model more complex, which may result in further instability.

It is also important to note that for each of the four algorithms discussed above, 
combining them with Prioritized Experience Replay resulted in 
better results both in terms of mean and SD of the accuracy. This is explained by the fact that in DQN, Double DQN, and Dueling DQN, during training, the experiences are sampled uniformly from the replay buffer. However, PER prioritizes these experiences such that those with a higher priority are sampled more often. Effectively, in a task such as ours, where most of the actions have a zero reward, this type of DQN favors transitions with a non-zero reward that are mainly the diagnosis actions. Sampling these actions more frequently, on which the accuracy is primarily based, leads the model to learn more and faster about them. 

\subsection{Quality of Diagnoses}
Regarding the comparison with SOTA classifiers, since we built our dataset using a decision tree, it was expected that the tree-based classifiers would perform extremely well. The performance of the two DQN Models was slightly lower than those of the tree-based methods and the neural network, while SVM had slightly lower performance.

In Table \ref{tab:classification_report}, we display the classification report of one particular experiment with the Dueling DQN-PER model, with an accuracy of 97.186, which is the closest accuracy to the mean. The report shows that most of the anemia classes exhibited a decent performance. However, while the \emph{No anemia} class had the best recall with a perfect score of 1, it also had the lowest precision at 0.92, meaning that several instances were being diagnosed as \emph{No anemia}, whereas they should not, based on their laboratory test results. Upon further inspection, we noted that all the misdiagnosed instances had values near the threshold value. In the \emph{No anemia} misclassifications, all the hemoglobin levels were close to the 13 and 12 g/dl thresholds for men and women, respectively (see Figure \ref{fig:tree}). In particular, the hemoglobin of men had a mean of 12.982 (SD=0.011), while the one of women had a mean of 11.931 (SD=0.042). However, it should be noted that from a clinical point of view, hemoglobin levels close to the normal threshold usually do not lead to an anemia diagnosis. 
Similarly, looking at \emph{Hemolytic anemia}, which had the lowest recall score at 0.94 (along with \emph{Aplastic anemia}), 35 of its instances were diagnosed with \emph{Vitamin B12/Folate deficiency anemia}. These instances had a mean MCV value of 99.765 (SD=0.170), thus close to the threshold of 100. The 11 other instances diagnosed with \emph{Anemia of chronic disease} had a mean MCV level of 80.197 (SD=0.152), close to the threshold of 80. And the other 30 instances with an \emph{Inconclusive diagnosis} all had values near the thresholds, but missing values for other features in the pathway leading to an inconclusive diagnosis. 

\begin{table}[h]
  \fontsize{7pt}{7pt}\selectfont
  \centering
  \begin{tabular}{c|c|c|c}
        \textbf{Class Name} & \textbf{Precision} & \textbf{Recall} & \textbf{F1-Score} \\
        \hline
        \textbf{Inconclusive} & \multirow{2}{*}{0.94} & \multirow{2}{*}{0.97} & \multirow{2}{*}{0.95} \\
        \textbf{diagnosis (1344)}& & \\
        \hline
        \textbf{Aplastic} & \multirow{2}{*}{1.00} & \multirow{2}{*}{0.94} & \multirow{2}{*}{0.97} \\
        \textbf{anemia (1806)}& & \\
        \hline
        \textbf{Hemolytic} & \multirow{2}{*}{1.00} & \multirow{2}{*}{0.94} & \multirow{2}{*}{0.97} \\
        \textbf{anemia (1805)}& & \\
        \hline
        \textbf{Iron deficiency} & \multirow{2}{*}{0.98} & \multirow{2}{*}{0.98} & \multirow{2}{*}{0.98} \\
        \textbf{anemia (1679)}& & \\
        \hline
        \textbf{Anemia of chronic} & \multirow{2}{*}{0.99} & \multirow{2}{*}{0.97} & \multirow{2}{*}{0.98} \\
        \textbf{disease (1772)}& & \\
        \hline
        \textbf{Unspecified} & \multirow{2}{*}{1.00} & \multirow{2}{*}{0.98} & \multirow{2}{*}{0.99} \\
        \textbf{anemia (1793)}& & \\
        \hline
        \textbf{Vitamin B12/Folate} & \multirow{2}{*}{0.96} & \multirow{2}{*}{0.98} & \multirow{2}{*}{0.97} \\
        \textbf{defic. anemia (1801)}& & \\
        \hline
        \textbf{No anemia (2000)} & 0.92 & 1.00 & 0.96\\
  \end{tabular}
  \caption{Classification report showing the detailed performance of the Dueling DQN-PER model. Classes of anemia diagnosis are reported with their respective support.}
  \label{tab:classification_report}
\end{table}

\subsection{DQN Model Robustness}
The DQN models also showed robustness to both noise and missing data, exhibiting a consistent performance throughout different levels of added noise and missing data. We believe that this is because DQN relies on neural networks that have been shown to be robust to noise in some cases \citep{goodfellow2016deep}. This is due to the fact that adding noise to the training input can act as a form of regularization by preventing overfitting and leading to better generalization of the neural network. Likewise, neural networks have also demonstrated their capability to handle missing data as they are able to learn the underlying representation of the data even with missing values. In this study, we remarked that introducing noise and missing data led to a 
decrease in performance of the \emph{Inconclusive diagnosis} and \emph{Iron deficiency anemia} classes (results not shown). This is attributed to the fact that some of the noisy instances and/or instances with missing data are being diagnosed with \emph{Inconclusive diagnosis} instead of their anemia class, which is to be expected. In the real world, in such cases, clinicians may need to use their own judgment. Additionally, many \emph{Anemia of Chronic Disease} instances are being diagnosed with \emph{Iron Deficiency Anemia} because they are on the same branch of the tree. 
Regarding the datasets that have both additional noise and missing data, SVM performs much lower than the rest, and while DT performs better, its performance is still significantly lower than the other models. This can be attributed to decision tree sensitivity to noise and missing data. In addition, they are prone to overfitting, such that a slight change in the training dataset may lead to a different tree altogether. Alternatively, Random Forest builds multiple decision trees during their training process and is therefore less likely to overfit than the DT, making it more impervious to noise and missing data.

Finally, using the same hyperparameters for the model, the DQN model was able to perform consistently as the training set size gradually decreased. The reason for this is the same hyperparameters were used for the experiments therefore a smaller dataset just meant that the same experiences were sampled more often. However, there is still a loss in the information gained from the data, and therefore the smaller the dataset, the more varied the results as shown by the confidence interval in Figure \ref{varying_sizes}.

\paragraph{Limitations}
One significant limitation of this study is the absence of experiments on a real-world dataset to evaluate our method. While the results from the synthetic dataset are very encouraging, and while we endeavor to reproduce imperfections of EHR data in our study, it is paramount that we test on real-world data. Additionally, the synthetic dataset used is not longitudinal. This initial study with anemia aimed to show the viability of the proposed method. Real-world EHR datasets often have a temporal dimension which will provide more insights into the progression of the pathways. Furthermore, with the availability of such data, we will be able to compare our approach with models capable of handling sequential data, such as Recurrent Neural Networks.

Another limitation is the sole consideration of anemia, whose diagnosis follows a decision tree. Other conditions are diagnosed based on guidelines that do not follow a decision tree. It would require novel experiments to know to what extent our approach may be generalized to other kinds of diagnoses such as Systemic Lupus Erythematosus (SLE), which has a more complex diagnosis process. For such a condition, the design of the state space as well as the reward function may be more intricate, and may also require more advanced techniques.

Moreover, training the RL agents took a significantly longer time and used considerably more computing resources than the SOTA classifiers as shown in Table \ref{tab:resources}. Also, the DQN model has many hyperparameters that would need to be further optimized. However, it should be noted that after model training, generating a diagnosis pathway for a test instance is trivial since the  policy has already been learned by the model (see Table \ref{tab:resources}). 

Finally, we did not evaluate Large Language Model-based methods, which have shown promising results on other tasks. We will take them into consideration in future work.

\section{Conclusion} \label{sec:conclusion}

In this work, we used DRL to generate personalized pathways for the diagnosis of anemia. We compared the performance of various DRL methods to each other and to state-of-the-art classifiers. We also tested the performance of DRL on noisy data, missing data, and on train sets of varying sizes. We demonstrated that our approach with DQN is suitable as it has a comparable performance with the SOTA classifiers with the added advantage of constructing personalized pathways for each patient. These pathways have the potential to guide the diagnosis of a new patient and to be aggregated to summarize possible pathways for a patient population. 
In future work, we aim to test our approach on different types of diagnoses, especially conditions whose diagnosis is not based on a decision tree. Additionally, since laboratory tests are usually ordered as a panel, this can be incorporated as well. It will also be interesting to test our approach on diagnoses that are not solely based on laboratory test results, but on multimodal data, potentially acquired over a significant  period of time. Finally, 
it will be crucial to assess our methods on real-world data.

\acks{This work is supported by the Inria CORDI-S Ph.D. program.}

\bibliography{bibliography}

\appendix
\renewcommand{\thetable}{\Alph{section}.\arabic{table}}
\renewcommand\thefigure{\thesection.\arabic{figure}}   

\section{Q-learning and its extensions}\label{apd:Q}
\setcounter{figure}{0} 
\setcounter{table}{0}

\textbf{Q-learning} \citep{watkins1992q} is an RL algorithm that outputs the best action to take in a given state based on the expected future reward of taking that action in that particular state. The expected future reward is named the Q-value of that state-action pair, 
noted $Q(s,a)$. At each time step, the agent selects an action following a policy $\pi$, and the goal is to find the optimal policy $\pi^*$ that maximizes the reward function. 
During model training, the Q-values are updated using the Bellman Equation as follows: 
\begin{align}
\label{eqn:q_learning}
Q(s_t, a_t) \leftarrow & Q(s_t, a_t) + \alpha[r_{t+1} +\gamma\ \underset{a}{\max}\  Q(s_{t+1}, a) -\nonumber\\
  &\quad Q(s_t, a_t)]
\end{align}

where $\alpha$ is the learning rate and $\gamma$ is the discount factor that determines the importance of future reward relative to immediate reward.


In our use case, since the problem has a large state space, we propose to use a \textbf{Deep Q-Network (DQN)} \citep{mnih2015human} which uses a neural network to approximate the Q-value function.
A DQN comprises two networks of similar architecture, \textit{i.e.}, the policy network, which interacts with the environment and learns the optimal policy, and the target network, which is used to define the target Q-value. At each time step, the policy network takes a state, $s_{t}$, as its input and outputs the Q-values for taking the different actions in that state. The weights of the target network are frozen and updated at a specified interval by copying the weights of the policy network. The DQN algorithm learns by minimizing the loss function shown in Equation \ref{eqn:dqn_loss_fn}, 
where $\theta$ and $\theta^-$ represent the weights of the policy and target networks, respectively. Additionally, at each time step, a record of the model's interaction with the environment (known as an experience) is stored in a memory buffer in the form $(s_t, a_t, r_{t+1}, s_{t+1})$.

\begin{equation}
    \label{eqn:dqn_loss_fn}
    L(\theta) = \mathop{\mathbb{E}}[(r_{t+1} + \gamma\ \underset{a}{\max}\  Q(s_{t+1}, a, \theta^-) - Q(s_t, a_t, \theta))^2]
\end{equation}

In order to improve DQN stability and performance, several extensions of the DQN algorithm have been developed which we use in this paper and we briefly describe below:

\textbf{Double DQN (DDQN)} \citep{van2016deep}: 
The action is selected using the policy network, while the target network estimates the value of that action unlike in the standard DQN where the same network is used for both tasks. The loss function is thus modified as follows:


\begin{align}
\label{eqn:ddqn_loss_fn}
 L(\theta) &= \mathop{\mathbb{E}}[(r_{t+1} + \gamma\ Q(s_{t+1}, \underset{a}{\arg\max}\ Q(s_{t+1}, a, \theta), \theta^-) \nonumber\\
  &\quad - Q(s_t, a_t, \theta))^2]
\end{align}

\textbf{Dueling DQN} \citep{wang2016dueling}: The Q-value function is split into two parts: a value function $V(s)$ that provides the value for being in that state, and an advantage function $A(s,a)$ that gives the advantage of the action $a$ in the state $s$, as compared to the other actions. The two functions are then combined to get the Q values as shown in Equation \ref{eqn:dueling_dqn}.
\begin{equation}
\label{eqn:dueling_dqn}
    Q(s,a) = V(s) + (A(s,a) - \frac{1}{|\mathcal{A}|} \underset{a}\sum A(s,a))
\end{equation}

where $|\mathcal{A}|$ is the total number of possible actions in that state.

\textbf{Prioritized Experience Replay (PER)} \citep{schaul2015prioritized}: 
Each experience in the buffer is assigned a priority such that experiences that have a higher priority are sampled more often during training.

\section{Dataset}\label{apd:dataset}
\setcounter{figure}{0} 
\setcounter{table}{0}
\subsection{Feature Inclusion}
The first step of our dataset construction was the definition of a set of features, associated or not, with anemia diagnosis. 
The hemoglobin level is the primary feature that is considered to determine whether a patient has anemia, therefore we included it in the dataset. Additionally, the normal level of hemoglobin varies between men and women, therefore gender was included too. We also included features from the decision tree of the standard guidelines for the diagnosis of anemia by \citet{bmj_anemia}. These include mean corpuscular volume (MCV), ferritin, reticulocyte count, segmented neutrophils and Total Iron Binding Capacity (TIBC). Additionally, we included features that are not in the tree, but that can be used to diagnose different types of anemia according to our discussions with a domain expert, namely, hematocrit, transferrin saturation (TSAT), red blood cells (RBC), serum iron and folate. Interestingly for our study, three of these features (hematocrit, TSAT and RBC) can be derived from other features and are accordingly correlated with features from our initial selection (\textit{e.g.}, hemoglobin and hematocrit). 
Furthermore, we added features that are not relevant to the diagnosis of anemia in order to observe their potential impact on the behavior of our model. These are creatinine, cholesterol, copper, ethanol, and glucose. Ultimately, a total of 17 features were included.

\subsection{Dataset construction}
The second step of the dataset construction was to build a dataset for 7 diagnosis classes. These are \emph{No anemia}, \emph{Vitamin B12/Folate deficiency anemia}, \emph{Unspecified anemia}, \emph{Anemia of chronic disease (ACD)}, \emph{Iron deficiency anemia (IDA)}, \emph{Hemolytic anemia}, and \emph{Aplastic anemia}. Each anemia class dataset was built based on the decision tree represented in Figure \ref{fig:tree}, which was manually constructed based on \citet{bmj_anemia} and \citet{short2013iron}.

\begin{figure*}[htbp]
\floatconts
  {fig:tree}
  {\caption{The decision tree used to create the dataset as adapted from \citet{bmj_anemia} and \citet{short2013iron}.}}
  {\includegraphics[width=\textwidth]{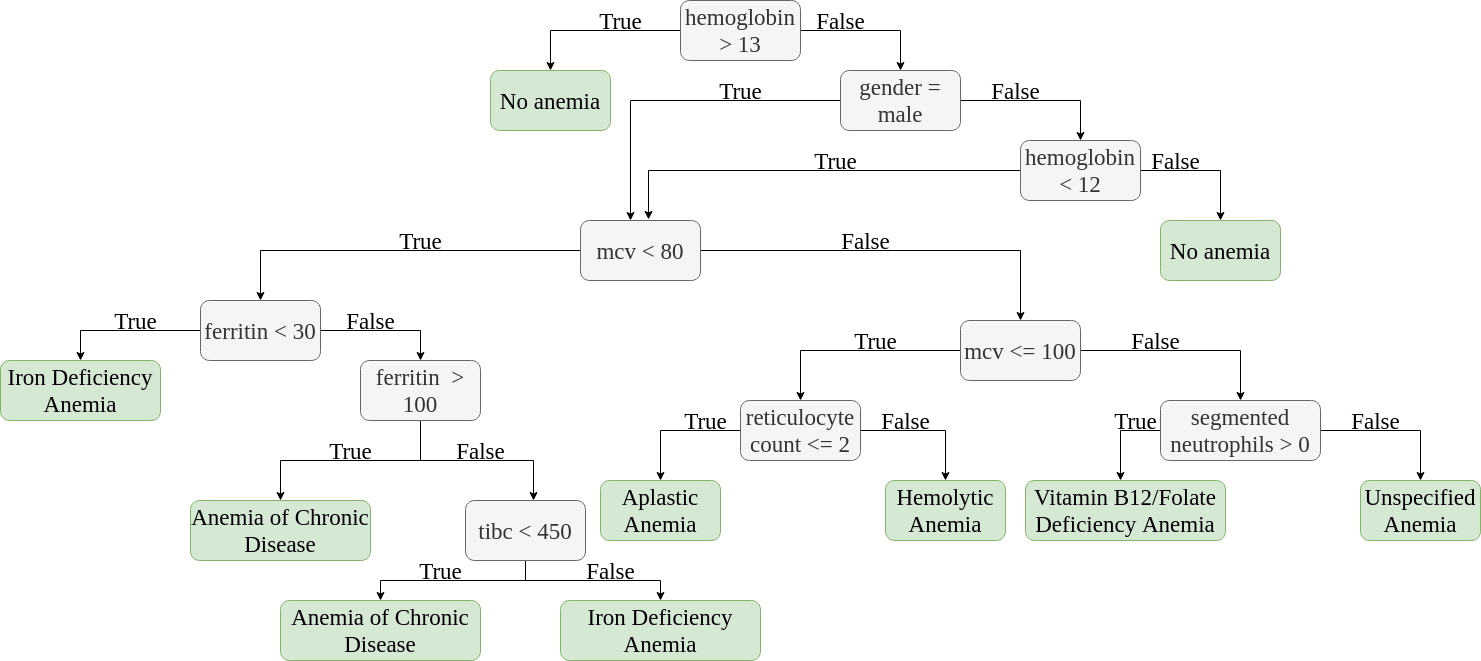}}
\end{figure*}

For each class, the values of the features were generated using a uniform probability distribution whose parameters (minimum and maximum values) were determined through manual reviewing of the medical literature and thresholds of the decision tree in Figure \ref{fig:tree}.  Features that are correlated with other features were derived using known equations, that is, hematocrit \citep{ht_formula}, TSAT \citep{tsat_formula} and RBC \citep{rbc_formula}.
10,000 instances were created for each of the diagnosis classes, which were then combined to create a single dataset of 70,000 instances. Next, an eighth diagnosis class, \emph{Inconclusive diagnosis} was created for when the model is not sure about the diagnosis of an instance. The \emph{Inconclusive diagnosis} instances were created out of the existing 70,000 instances by randomly selecting and removing 10\% of the non-missing values of each feature (except hemoglobin, gender and MCV, which are necessary to the diagnosis of almost all the anemia classes). 

\begin{figure}[htbp]
\floatconts
  {fig:class_distribution}
  {\caption{Number of patients per anemia class.}}
  {\includegraphics[width=\linewidth]{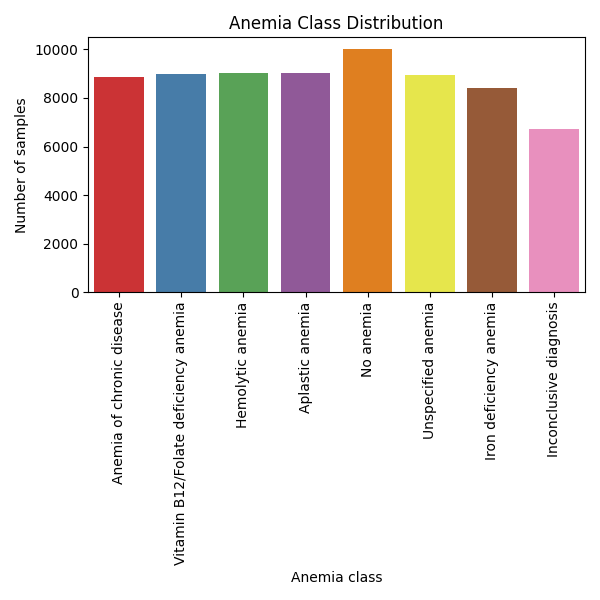}}
\end{figure}

\begin{figure}[htbp]
\floatconts
  {fig:missing_values}
  {\caption{Number of present vs missing values for each feature in the dataset.}}
  {\includegraphics[width=\linewidth]{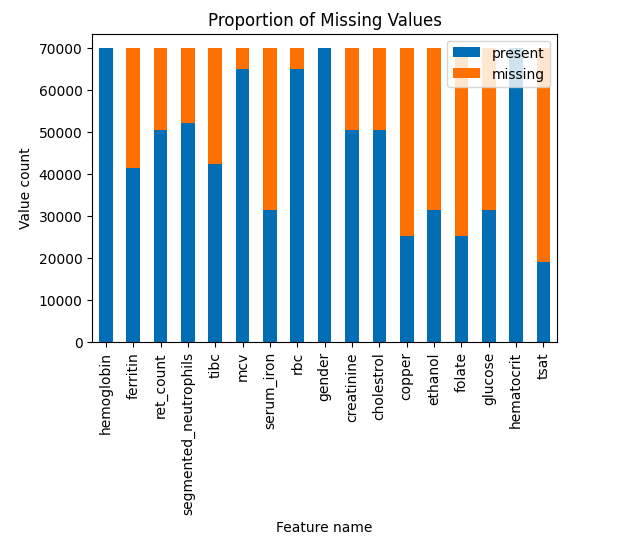}}
\end{figure}

\begin{table*}
    \fontsize{7pt}{7pt}\selectfont
    \centering
    \floatconts
    {tab:statistics}
    {\caption{Summary descriptive statistics of the dataset showing the mean and interquartile range (in parentheses) of the features. Gender, which is a binary variable, is described using the sample number and the percentage.}}
    {
    \begin{tabular}[]{@{}|l|l|l|l|@{}}
  \hline
    \bfseries Feature & \bfseries All Classes & \bfseries No anemia & \bfseries \makecell{Vitamin B12/Folate \\ deficiency anemia} \\ \hline
    \textbf{Hemoglobin} &  10.239 (8.067, 12.102) & 14.570 (13.281, 15.848) & 9.510 (7.771, 11.274) \\
    \textbf{Ferritin} &  209.968 (69.870, 343.337) & 251.436 (128.145, 373.501) & 251.775 (126.676, 374.486) \\
    \textbf{Reticulocyte count} & 2.821 (1.270, 4.342) & 2.981 (1.520, 4.423) & 3.024 (1.526, 4.564)\\
     \textbf{Segmented neutrophils} & 2.930 (0.762, 4.898) & 3.532 (1.824, 5.266) & 3.516 (1.791, 5.22)\\
     \textbf{TIBC} & 334.276 (222.108, 457.942) & 310.821 (206.744, 415.958) &  306.678 (202.380, 410.634)  \\
     \textbf{MCV} & 89.998 (78.918, 101.093) & 90.029 (82.631, 97.598) & 102.504 (101.225, 103.794) \\
     \textbf{Serum iron} & 135.030 (77.793, 192.645) & 134.959 (76.979, 192.764) & 136.613 (80.087, 193.071) \\
     \textbf{RBC} & 3.348 (2.641, 3.936) & 4.899 (4.379, 5.356) & 2.784 (2.274, 3.298)\\
     \textbf{Gender} & & & \\
        \hspace{5mm}\textbf{Male} & 38268 (54.67\%) & 4048 (40.48\%) & 5108 (56.73\%) \\
        \hspace{5mm}\textbf{Female} & 31732 (45.33\%) & 5952 (59.52\%) & 3896 (43.27\%)\\
     \textbf{Creatinine} & 1.103 (0.651, 1.552) & 1.119 (0.670, 1.566) & 1.102 (0.655, 1.540)\\
     \textbf{Cholestrol} & 74.878 (37.388, 112.244) & 74.037 (36.257, 111.019) & 75.020 (37.099, 112.819) \\
     \textbf{Copper} & 80.095 (55.182, 105.245) & 79.510 (54.965, 104.816) & 80.001 (55.014, 105.150)  \\
     \textbf{Ethanol} & 39.887 (19.876, 59.749) & 40.256 (19.769, 60.865) & 39.445 (18.622, 59.590) \\
     \textbf{Folate} & 15.262 (7.832, 22.715) & 15.054 (7.843, 22.174) & 15.400 (8.099, 22.976) \\
     \textbf{Glucose} & 90.039 (65.128, 115.077) & 90.918 (66.287, 115.892) & 90.021 (64.290, 116.194)\\
     \textbf{Hematocrit} & 30.716 (24.201, 36.306) & 43.709 (39.843, 47.544) & 28.530 (23.313, 33.821) \\
     \textbf{TSAT} & 49.601 (23.103, 62.608) & 52.553 (24.626, 67.235) & 53.935 (25.712, 68.534) \\
  \hline
\end{tabular}

\bigskip

\begin{tabular}[]{@{}|l|l|l|l|@{}}
  \hline
    \bfseries Feature & \bfseries Unspecified anemia & \bfseries \makecell{Anemia of \\ chronic disease} & \bfseries Iron deficiency anemia \\\hline
         \textbf{Hemoglobin} & 9.534 (7.769, 11.301) & 9.514 (7.751, 11.254) & 9.539 (7.795, 11.276) \\
         \textbf{Ferritin} & 250.757 (123.567, 375.763) & 268.551 (152.371, 384.475) & 48.654 (22.742, 74.091) \\
         \textbf{Reticulocyte count} & 3.006 (1.546, 4.459) & 2.957 (1.458, 4.457) & 2.975 (1.441, 4.502) \\
         \textbf{Segmented neutrophils} &  0.000 (0.000, 0.000) & 3.580 (1.848, 5.284) & 3.582 (1.854, 5.330)\\
         \textbf{TIBC} & 311.332 (208.894, 414.291) & 301.558 (199.155, 402.335) & 452.223 (458.074, 499.101) \\
         \textbf{MCV} & 102.525 (101.274, 103.774) & 77.472 (76.206, 194.049) & 77.527 (76.299, 78.796)\\
         \textbf{Serum iron} & 134.970 (77.499, 192.216) & 135.313 (77.093, 194.049) &  135.625 (77.915, 193.272)\\
         \textbf{RBC} & 2.790 (2.274, 3.305) & 3.685 (3.008, 4.361)  & 3.693 (3.013, 4.371) \\
         \textbf{Gender}  & & & \\
            \hspace{5mm}\textbf{Male} & 5175 (57.73\%) & 5067 (57.20\%) & 4765 (56.74\%)\\
            \hspace{5mm}\textbf{Female} & 3789 (42.27\%) & 3792 (42.8\%) & 3633 (43.26\%) \\
         \textbf{Creatinine} & 1.096 (0.643, 1.554) & 1.098 (0.643, 1.544) & 1.103 (0.643, 1.558) \\
         \textbf{Cholestrol} & 75.734 (38.838, 113.730) & 74.994 (37.212, 112.230) & 74.700 (37.507, 112.034) \\
         \textbf{Copper} & 80.465 (56.729, 105.069) & 80.081 (54.461, 104.850) & 79.764 (54.906, 104.545)\\
         \textbf{Ethanol} & 39.500 (19.102, 22.784) & 39.784 (19.802, 59.703) & 39.501 (20.018, 58.701) \\
         \textbf{Folate} & 15.284 (7.919, 22.784) & 15.131 (7.769, 22.650) & 15.462 (7.948, 22.976)\\
         \textbf{Glucose} & 89.457(65.322, 113.808) & 90.131 (66.091, 115.144) & 90.153 (65.518, 114.407)\\
         \textbf{Hematocrit} & 28.602 (23.307, 33.902) & 28.541 (23.253, 33.761) & 28.617 (23.386, 33.828)\\
        \textbf{TSAT} & 52.689 (24.323, 67.385) & 54.538 (24.730, 71.416) & 32.675 (17.377, 42.591)\\\hline
\end{tabular}

\bigskip
\begin{tabular}[]{@{}|l|l|l|l|@{}}
  \hline
    \bfseries Feature & \bfseries Hemolytic anemia & \bfseries Aplastic anemia & \bfseries \makecell{Inconclusive \\ diagnosis }\\\hline
    \textbf{Hemoglobin} & 9.510 (7.741, 11.262) & 9.518 (7.801, 11.250) & 9.486 (7.764, 11.224)\\
    \textbf{Ferritin} & 251.082 (123.988, 380.350) & 246.611 (119.831, 374.303) & 195.158 (67.362, 319.364)\\
    \textbf{Reticulocyte count} & 4.049 (3.079, 5.007) & 1.005 (0.514, 1.500) & 2.988 (1.437, 4.514)\\
     \textbf{Segmented neutrophils} & 3.553 (1.820, 5.271) & 3.517 (1.789, 5.189) & 3.478 (1.745, 5.214)\\
     \textbf{TIBC} & 310.197 (203.256, 418.237) & 310.744 (205.529, 414.594) & 338.282 (225.384, 461.215)\\
     \textbf{MCV} & 89.944 (84.920, 94.938) & 89.985 (85.003, 95.013) & 88.696 (78.100, 100.800)\\
     \textbf{Serum iron} & 133.068 (76.467, 191.047) & 135.668 (79.215, 192.992) & 133.704 (76.167, 191.807)\\
     \textbf{RBC} & 3.185 (2.580, 3.761)  & 3.186 (2.596, 3.747)  & 3.257 (2.628, 3.809)\\
     \textbf{Gender} & & & \\
        \hspace{5mm}\textbf{Male} & 5166 (57.24\%) & 5084 (56.31\%) & 3855 (57.36\%)\\
        \hspace{5mm}\textbf{Female} & 3859 (42.76\%) & 3945 (43.69\%) & 2866 (42.64\%)\\
     \textbf{Creatinine} & 1.098 (0.656, 1.538) & 1.100 (0.646, 1.551) & 1.108 (0.646, 1.563)\\
     \textbf{Cholestrol} & 74.968 (37.216, 112.418) & 74.990 (37.919, 112.314) & 74.602 (37.416, 111.479)\\
     \textbf{Copper} & 80.348 (55.201, 105.698) & 80.502 (55.372, 106.089) & 80.166 (55.087, 105.649)\\
     \textbf{Ethanol} & 40.220 (20.430, 60.389) & 40.525 (20.545, 60.523) & 39.744 (21.113, 58.165)\\
     \textbf{Folate} & 15.396 (7.854, 22.746) & 15.309 (7.731, 22.812) & 15.045 (7.521, 22.613)\\
     \textbf{Glucose} & 90.140 (65.137, 115.555) & 89.852 (64.575, 114.951) & 89.348 (63.492, 113.911)\\
     \textbf{Hematocrit} & 28.530 (23.223, 33.785) & 28.554 (23.402, 33.751) & 28.458 (23.292, 33.671)\\
     \textbf{TSAT} & 53.800 (24.902, 69.688) & 53.292 (26.181, 69.297) & 48.753 (21.476, 63.018)\\\hline
\end{tabular}
}
    
\end{table*}

\begin{table*}[htbp]
\centering
\floatconts
  {tab:sample}
  {\caption{An instance in the dataset.}}%
  {%
    \begin{tabular}[]{@{}|l|l|@{}}
    \hline
         \bfseries Feature & \bfseries Value \\\hline
         \textbf{Hemoglobin} & 9.007012 \\
         \textbf{Ferritin} &  - \\
         \textbf{Reticulocyte count} & - \\
         \textbf{Segmented neutrophils} & 3.519565 \\
         \textbf{TIBC} & 440.499323 \\
         \textbf{MCV} & 103.442762 \\
         \textbf{Serum iron} & 59.017997 \\
         \textbf{RBC} & 2.612173 \\
         \textbf{Gender} & Male \\
         \textbf{Creatinine} & 0.650757 \\
         \textbf{Cholestrol} & 114.794964 \\
         \textbf{Copper} & 112.308159 \\
         \textbf{Ethanol} & 25.612786 \\
         \textbf{Folate} & 5.969710 \\
         \textbf{Glucose} & 116.026042 \\
         \textbf{Hematocrit} & 27.021037 \\
         \textbf{TSAT} & 13.397977 \\
         \textbf{label} & Vitamin B12/Folate deficiency anemia\\\hline
    \end{tabular}
  }
\end{table*}

\subsection{Simulating Imperfect Data}\label{sec:imperfect}
To compare the robustness of various approaches to imperfect data, we artificially introduced different levels of noisiness and missingness to our training dataset.

For missingness, a percentage of the values for each feature, excluding hemoglobin and gender, were randomly replaced with missing values. Hemoglobin and gender were excepted because a patient's hemoglobin level is key to the diagnosis of anemia and the normal levels of hemoglobin vary between men and women. 

Since our original dataset perfectly follows a decision tree, we simulated noisiness using the following procedure. For each anemia class, except \emph{No anemia} and \emph{Inconclusive diagnosis}, a specified fraction of the values in each of the features in its branch of the tree in Figure \ref{fig:tree} were replaced by another value generated from a normal distribution, $N(\mu, \sigma^2)$, where the mean, $\mu$, is the threshold in a node with that feature in the tree, and the standard deviation, $\sigma$, was defined by us.

For example, using hemolytic anemia, the features in its tree branch, that is, the features used to diagnose a patient with it are \emph{hemoglobin}, \emph{MCV} and \emph{reticulocyte count} as shown in Figure \ref{fig:tree}. Therefore, for a noise level of 0.2, we replaced 20\% of the \textit{reticulocyte count} values using a normal distribution function $N(2, 0.2)$ where $\mu$ = 2.0 since this is the threshold in the decision tree. Similarly, for \textit{MCV}, 10\% of its feature values were replaced using a normal distribution $N(80, 2)$ while another 10\% was replaced using a normal distribution function $N(100, 2)$. We gradually increased the noise level fraction in order to analyze its effect on the performance of our model and the pathways it creates. Additionally, for all the noise levels, 10\% of the anemic instances were randomly labeled as \textit{No anemia}. 

We further created datasets with both noisy and missing data. Using a training dataset with a noisiness level of 0.2 as described above, we added missing data at different levels to this dataset using the same procedure used to create the datasets with missing data.

\section{Model Hyperparameters}\label{apd:params}
\setcounter{figure}{0} 
\setcounter{table}{0} 

Table \ref{tab:hyperparams} shows the values of the hyperparameters used for the DQN model in this study.

\begin{table*}
\floatconts
  {tab:hyperparams}
  {\caption{DQN hyperparameter values.}}%
  {%
    \begin{tabular}{|l|l|}
    \hline
    \abovestrut{2.2ex}\bfseries Hyperparameter & \bfseries Value \\\hline
        Buffer size & 1000000 \\ 
        Learning rate & 0.0001 \\
        Target network update frequency & 10000 \\ 
        Learning starts & 50000 \\
        Final epsilon value & 0.05 \\
        Discount factor & 0.99 \\
    \belowstrut{0.2ex}Train frequency & 4\\\hline
    \end{tabular}
  }
\end{table*}

\section{Sample Learned Pathways}\label{apd:pathway}
\setcounter{figure}{0} 
\setcounter{table}{0}

\begin{figure*}[h]
\floatconts
  {fig:side_by_side}
  {\caption{A side-by-side illustration of (\textit{a}) a branch of the decision tree in Figure \ref{fig:tree} and (\textit{b}) the corresponding pathways learned by the model for \emph{No anemia}, \emph{Hemolytic anemia} and \emph{Unspecified anemia}.}}
  {%
    \subfigure[]{\label{fig:tree_path1}%
      \includegraphics[width=0.49\linewidth]{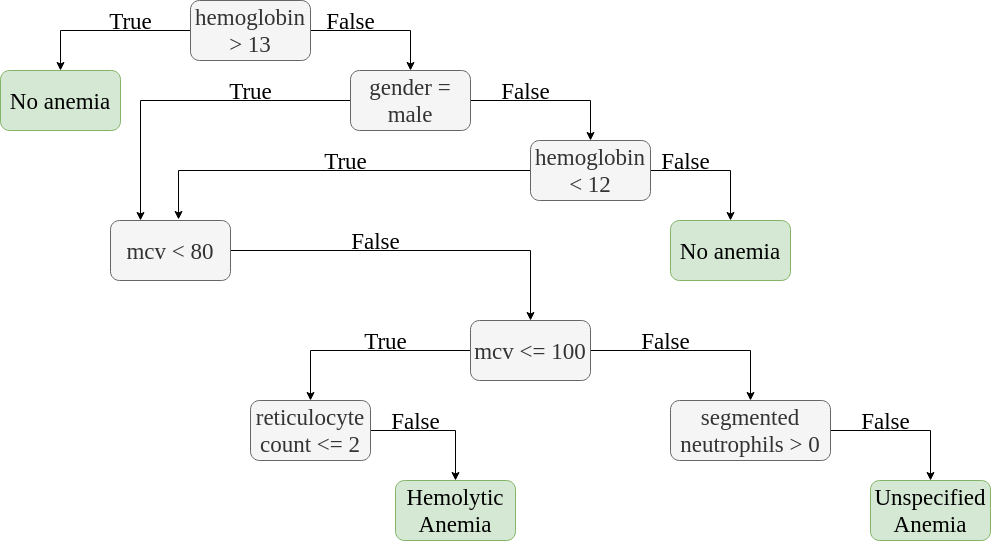}}%
    \hfill
    \subfigure[]{\label{fig:pathway_side}%
      \includegraphics[width=0.49\linewidth]{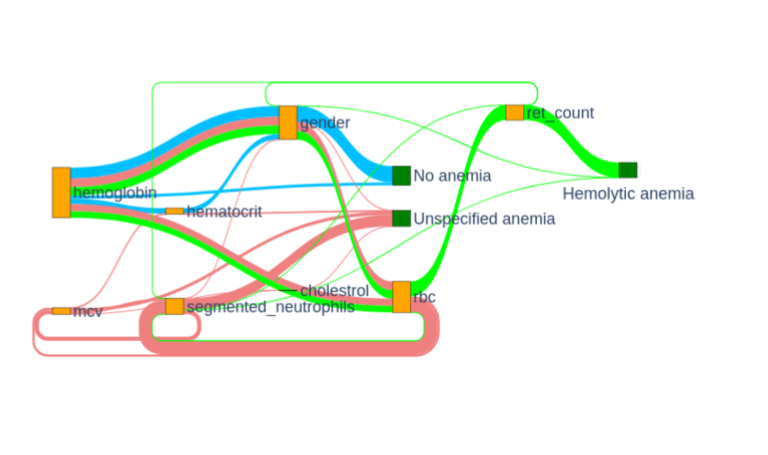}}
  }
\end{figure*}


\begin{figure*}[h]
\floatconts
  {fig:pathway2}
  {\caption{The clinical diagnosis pathways for \emph{ACD} and \emph{Aplastic anemia} as learned by the agent.}}
  {\includegraphics[scale=0.5]{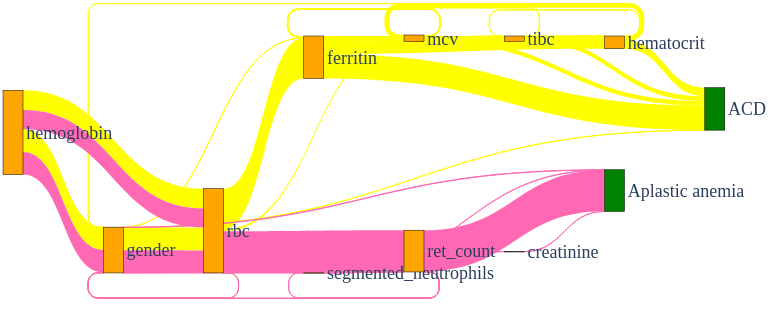}}
\end{figure*}

Figures \ref{fig:side_by_side} and \ref{fig:pathway2} provide examples of pathways generated with our approach. In particular, Figure \ref{fig:side_by_side} is a side-by-side illustration of a branch of the decision tree used to label the dataset and a Sankey diagram showing the pathways learned by the Dueling DQN-PER model for the diagnosis of solely \emph{No anemia}, \emph{Unspecified anemia} and \emph{Hemolytic anemia}, which are colored \emph{blue}, \emph{coral} and \emph{light green}, respectively. The value actions are represented by the orange nodes, while the diagnosis actions are the dark green nodes. In addition, the size of each flow corresponds to its support, \textit{i.e.}, the number of patients that have it in their pathway.
Figure \ref{fig:pathway2} shows the pathways learned by the model for the diagnosis of \emph{Anemia of Chronic Disease} and \emph{Aplastic anemia} which are colored \emph{yellow} and \emph{pink} in that order.
One could enrich these visualizations with the threshold values that lead to one path or another.

\section{Computing Time}\label{apd:resources}
\setcounter{figure}{0} 
\setcounter{table}{0} 

\begin{table*}[!p]
\centering
\floatconts
  {tab:resources}
  {\caption{The computing time to train the model and to generate a diagnosis (pathway) for a single test instance.}}%
  {%
    \begin{tabular}{|p{3.5cm}|p{4.4cm}|p{4.4cm}|}
    \hline
    \bfseries Model & \bfseries Training time & \bfseries Testing time\\
    \hline
         Decision Tree & 299 ms ± 9.72 ms  & 18.6 µs ± 453 ns \\ 
         Random Forest & 5.85 s ± 8.56 ms & 3.27 ms ± 156 µs \\
         SVM & 25.8 s ± 437 ms & 166 µs ± 8.33 µs \\ 
         FFNN & 2min 36s ± 35 s & 645 µs ± 28.6 µs \\ 
         Dueling DQN-PER & 1h 42 min 14s ± 19 min 19s & 722 µs ± 35.1 µs \\ 
         Dueling DDQN-PER & 2h 12min 45s ± 7min 36s & 759 µs ± 56.1 µs \\
    \hline
    \end{tabular}
  }
\end{table*}

In Table \ref{tab:resources}, the time taken to train each model (training time), as well as, the time taken to generate a diagnosis pathway or a diagnosis (testing time) for the DQN models and the SOTA models respectively for a single instance in the test dataset are shown.

\end{document}